# Morphology-Aware Multimodal Representation Learning for Insect Phylogenetic Reconstruction

Zixuan Liu, Kaijie Yu, Chun He, Xiaoxu Cai, Xinhai Ye, Haishuai Wang, Gongyin Ye, Jiajun Bu

***Abstract*—Morphological traits provide important evidence for phylogenetic reconstruction and evolutionary relationship analysis. Recent image-based approaches have introduced deep learning, particularly convolutional models, to derive morphological features from specimen images, but these methods generally rely on single-modality visual representations and do not explicitly incorporate morphological semantics. This study proposes a morphology-aware multimodal alignment framework for insect phylogenetic reconstruction. The framework combines specimen images with curated morphological descriptions by adapting a vision transformer through parameter-efficient fine-tuning and supervised contrastive learning, followed by image-text alignment in a shared latent space. The learned image embeddings are then used as continuous traits for Bayesian phylogenetic reconstruction. On the public Rove-Tree-11 dataset, comparative and ablation experiments across multiple visual backbones and feature adaptation strategies demonstrate that multimodal alignment improves topological agreement with the reference phylogeny. The results indicate that the proposed framework can derive morphology-aware visual traits for computational phylogenetic reconstruction.**

## I. Introduction

MORPHOLOGICAL traits provide an important source of evidence for reconstructing evolutionary relationships [1]. Morphology remains especially useful for taxa and specimens where molecular sampling is incomplete, unavailable, or difficult to obtain [2]. It also provides a complementary view of organismal evolution in the genomic era [3]. In this context, computational phylogenetic reconstruction requires learned representations that capture morphology-relevant variation while remaining suitable for downstream statistical tree inference. [4].

Traditional morphology-based inference relies on manually selected anatomical characters or quantitative measurements. These representations are interpretable, but their construction requires expert knowledge and is difficult to scale to large digitized collections. Recent image-based studies have begun to use learned visual traits from specimen images, demonstrating the potential of computational feature extraction for evolutionary analysis. However, most image-derived approaches still treat morphology primarily as visual information, or use image-derived traits as one evidence source in broader phylogenetic analyses, without explicitly aligning visual features with curated morphological semantics.

At the same time, advances in convolutional and transformer-based architectures, together with self-supervised learning, have substantially improved the ability to learn transferable visual representations from large-scale image collections. These models provide useful feature extractors, but their generic embeddings are not designed specifically for morphology-aware phylogenetic trait construction. A key challenge is therefore to adapt pretrained visual representations so that they preserve morphology-related variation useful for downstream tree reconstruction while remaining connected to explicit domain semantics.

Multimodal learning provides a mechanism for incorporating semantic supervision into learned representations. Image–text contrastive learning has demonstrated that natural-language supervision can organize visual embedding spaces [5], while related multimodal strategies have been applied to molecular representation learning and other computational biology tasks. However, these studies primarily focus on molecular structures, graphs, sequences, or heterogeneous feature fusion. To the best of our knowledge, no existing framework uses image–text alignment with curated morphological descriptions to guide the learning of continuous visual traits for insect phylogenetic reconstruction.

In this study, we propose a morphology-aware multimodal alignment framework for insect phylogenetic reconstruction. The framework adapts a self-supervised vision transformer to specimen images through parameter-efficient fine-tuning and supervised contrastive learning, encodes curated genus-level morphology descriptions as a fixed text representation, aligns image and text features in a shared latent space, and exports the learned image embeddings as continuous traits for Bayesian phylogenetic reconstruction. The main contributions are as follows:

- We propose an image-text alignment framework for morphology-aware insect representation learning;
- We design a transformer-based visual adaptation

*(Corresponding author: Xiaoxu Cai.)*

Zixuan Liu, Xiaoxu Cai, Haishuai Wang, Jiajun Bu are with the Zhejiang Key Lab of Accessible Perception and Intelligent Systems, Zhejiang University, Hangzhou 310027, China. (e-mail: liuzx068@zju.edu.cn)

Kaijie Yu, Chun He, Gongyin Ye are with the State Key Laboratory of Rice Biology and Breeding & Ministry of Agricultural and Rural Affairs Key Laboratory of Molecular Biology of Crop Pathogens and Insects, Institute of Insect Sciences, Zhejiang University, Hangzhou 310058, China.

Xiaoxu Cai is with the Rural Development Academy, Zhejiang University, Hangzhou 310058, China.

Xinhai Ye is with the College of Advanced Agriculture Science, Zhejiang A&F University, Hangzhou 311300, China.

The source code, data, and implementation details are publicly available at https://github.com/Camille09090/Multimodal-Phylogeny-/tree/main.

module that combines LoRA and supervised contrastive learning;
- We construct curated morphology text supervision for cross-modal alignment;
- We evaluate the framework on Rove-Tree-11 through model comparison, ablation-style analysis, topology metrics, and attention-based visual analytics.

The remainder of this paper is organized as follows. Section II reviews related work. Section III presents the methodology. Section IV describes the datasets, experiment settings, and baselines. Section V reports the results. Section VI concludes the paper.

## II. RELATED WORK

### *A. Morphological Representations for Phylogenetic Inference*

Morphological characters have long been used for phylogenetic reconstruction [1]. Discrete anatomical characters remain important for morphology-based systematics [2]. Continuous quantitative traits can also be incorporated into likelihood-based phylogenetic models [6]. Brownian-motion models provide a common statistical assumption for continuous-character evolution [7]. Despite their biological interpretability, manually coded morphological matrices are labor-intensive and can vary with character definition and taxon sampling [3].

Image-derived morphological traits provide a scalable alternative. Rove-Tree-11 introduced a hierarchically structured rove beetle image dataset for deep metric learning research [8]. Building on this dataset, Hunt et al. [9] extracted morphological traits from digitized specimen images using deep learning and integrated them with molecular data for total-evidence phylogenetic reconstruction, demonstrating the potential of image-derived representations for large-scale phylogenetic analysis. More broadly, deep learning on butterfly phenotypes has shown that image embeddings can capture biologically meaningful phenotypic variation [10]. These studies motivate learned morphology for evolutionary analysis, while also highlighting an open space for semantic alignment between visual traits and explicit morphological descriptions.

### *B. Visual Representation Learning*

Image representation learning has rapidly evolved from convolutional models to transformer-based architectures. Vision transformers model images as patch sequences and provide a flexible architecture for large-scale visual learning [11]. DINO demonstrates that self-supervised vision transformers can produce transferable visual representations without dense manual annotation [12]. DINOv2 extends this direction by training robust all-purpose visual features at larger scale [13]. EVA-02 represents large-scale visual representation learning [14]. BEiT v2 uses masked image modeling with vector-quantized visual tokenizers [15]. ConvNeXt V2 combines modern convolutional design with masked autoencoder pretraining [16].

For morphology-based phylogenetic reconstruction, these models are relevant not as generic classifiers, but as candidate feature extractors. The central issue is whether their embeddings preserve morphological structure that is informative for tree reconstruction. This study therefore compares multiple visual backbones and then adapts DINOv2 as the main image encoder for multimodal alignment.

### *C. Multimodal Learning in Computational Biology*

Multimodal learning aims to combine complementary information from different data views. CLIP-style image-text learning aligns visual and textual representations through contrastive objectives and has become a general paradigm for language-supervised visual representation learning [5]. BioCLIP extends this direction to biological data and provides a vision-language model for the tree of life [17].

In computational biology, multimodal models have been used to combine molecular structures, graphs, sequences, natural-language descriptions, fingerprints, and learned semantic embeddings. MoleculeSTM jointly learns molecule structures and text descriptions for retrieval and editing tasks [18]. MoMu bridges molecular graphs and natural language through multimodal pretraining [19]. Multimodal contrastive learning has also been used to transfer molecular property information from text to graph representations [20]. Recent studies have applied multimodal contrastive learning to molecular property prediction [21] and attention-based multimodal fusion to protein inhibitor prediction [22]. This study investigates image – text alignment for learning morphology-derived continuous traits in insect phylogenetic reconstruction.

## III. METHODOLOGY

### *A. Overview*

The proposed framework consists of an image encoder, a text encoder, an alignment objective, and a phylogenetic reconstruction module. Fig. 1 illustrates the pipeline. Specimen images are processed by a DINOv2-based image encoder, while curated genus-level morphological descriptions are encoded into a fixed text feature bank. Image and text features are projected into a shared latent space and optimized by supervised contrastive learning and image-text alignment. After training, the image-side CLS embeddings are aggregated at genus level and used as continuous traits for Bayesian phylogenetic reconstruction.

### *B. Image Encoder*

The image encoder is based on DINOv2, a self-supervised visual model that produces robust visual features [13]. DINOv2 uses the vision transformer architecture introduced for image recognition [11] and builds on the self-supervised DINO training paradigm [12]. Each specimen image is encoded as patch tokens together with a CLS token, and the final CLS representation is used as the global image embedding.

To adapt the pretrained visual representation to insect morphology and phylogenetic reconstruction without full model training, low-rank adaptation (LoRA) modules are inserted into the image encoder [23]. Specifically, LoRA is applied to the self-attention projection layers of the DINOv2 Vision Transformer backbone. The pretrained weights of the linear projection layers are kept frozen, while trainable low-rank update matrices are introduced to model task-specific adaptations. In implementation, LoRA modules are inserted into the query and value projection matrices of all transformer blocks, while the patch embedding layer and all original backbone parameters remain fixed. This design enables efficient adaptation of attention responses to insect morphological patterns with a minimal number of trainable parameters, thereby reducing overfitting on the genus-level dataset.

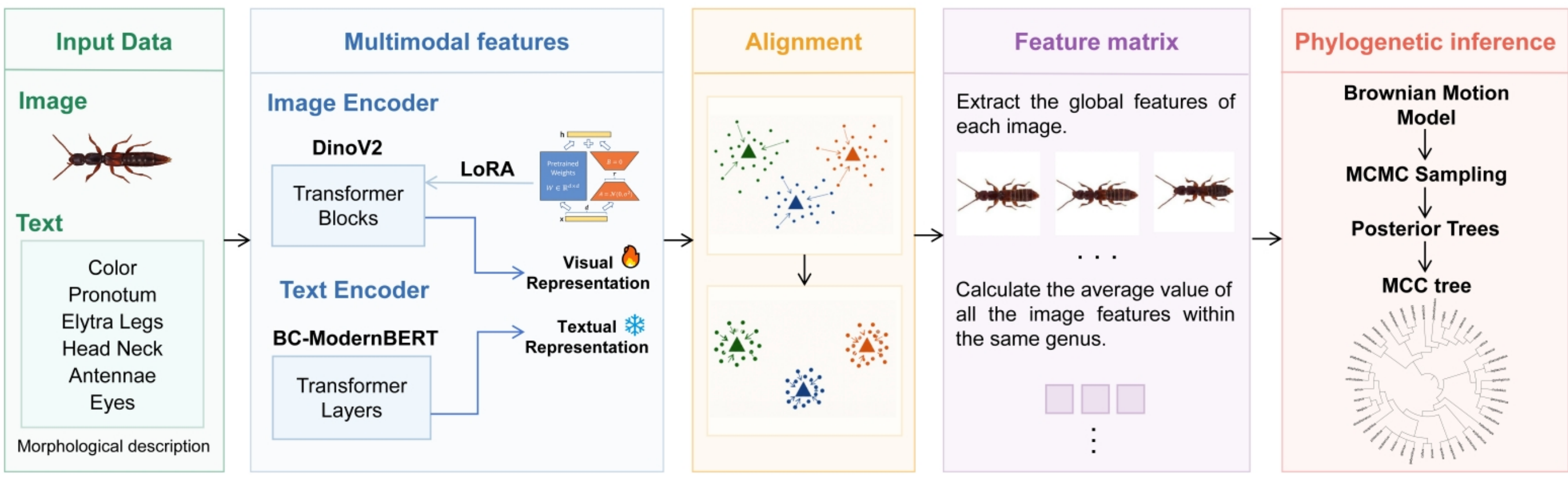

**Fig. 1.** Workflow of the proposed morphology-aware multimodal alignment framework.

Supervised contrastive learning is used to encourage embeddings from the same genus to be close while separating embeddings from different genera [24]. The supervised contrastive loss is defined as:

$$L_{\text{supcon}} = \frac{1}{N}\sum_{i=1}^{N}\left[log\sum_{j\neq i} exp\left(\frac{z_i \cdot z_j}{\tau}\right) - \frac{1}{|P(i)|}\sum_{p\in P(i)} \frac{z_i \cdot z_p}{\tau}\right] \tag{1}$$

where $z_i$ denotes the normalized embedding of the $i$-th sample, $\tau$ is the temperature coefficient, and $P(i)$ represents the set of positive samples that share the same class label as sample $i$.

After training, the CLS embeddings are extracted as global visual morphology features. Genus-level representations are obtained by averaging image features within each genus. These vectors are used as continuous traits in the downstream reconstruction module.

## C. Text Encoder

For the text branch, each curated genus-level morphological description is tokenized and encoded with BioClinical-ModernBERT-base, a domain-adapted transformer encoder for biomedical and clinical text [25]. The encoder first maps the tokenized description into contextual token representations, and a sequence-level embedding is then extracted as the semantic representation of the corresponding genus. This embedding summarizes the curated morphological descriptors rather than raw dataset metadata.

Because the text descriptions are defined at the genus level, the encoded vectors form a genus-level text embedding bank. During multimodal training, this bank is kept frozen and the text vector is retrieved according to the genus label of each image mini-batch. The retrieved text embedding is then passed through a text projection head before being aligned with the image representation in the shared latent space.

This design separates text encoding from image adaptation: the language model provides stable semantic anchors, while the image encoder and projection heads are optimized to match visual morphology to the curated textual descriptors.

## D. Alignment

Image and text representations are mapped into a shared latent space by two projection heads. For each mini-batch, the text feature corresponding to the genus label is retrieved from the fixed text bank.

The overall optimization objective is formulated as:

$$L = L_{\text{supcon}} + \lambda L_{\text{align}} \tag{2}$$

where $L_{\text{supcon}}$ denotes the supervised contrastive loss applied to image embeddings, and $L_{\text{align}}$ represents the image – text alignment loss. The hyperparameter λ controls the contribution of the alignment constraint.

The image-text alignment loss is defined as:

$$L_{\text{align}} = \frac{1}{C}\sum_{c=1}^{C}\left(1 - \frac{\bar{z}_c^{\text{img}} \cdot z_c^{\text{txt}}}{\|\bar{z}_c^{\text{img}}\| \, \|z_c^{\text{txt}}\|}\right) \tag{3}$$

where C denotes the total number of classes, $\bar{z}_c^{\text{img}}$ is the mean image feature representation of class c, and $z_c^{\text{txt}}$ is the corresponding text feature representation of class C.

During training, the LoRA modules and projection heads are updated, while the text feature bank remains frozen. The alignment objective acts as semantic regularization, encouraging image embeddings to preserve morphology-related information represented in the curated descriptions.

## E. Phylogenetic Reconstruction

The learned image embeddings are converted into a continuous-character matrix for phylogenetic inference. Each genus is represented by a 768-dimensional vector obtained from the adapted image encoder. The feature matrix is exported in NEXUS format and treated as continuous morphological traits.

Bayesian phylogenetic inference is performed in RevBayes [26]. The learned embeddings are analyzed with a Brownian-motion model for continuous-character evolution [7]. This follows the likelihood framework for evolutionary trees from continuous traits [6]. The analysis jointly estimates tree topology, branch lengths, and evolutionary rate parameters. Tree topology is explored using nearest-neighbor interchange and subtree pruning and regrafting moves, and posterior samples are summarized as a maximum clade credibility tree.

### *F. Evaluation Metrics*

Topological agreement is evaluated using normalized Robinson-Foulds (RF) distance [27] and normalized alignment score. Let $\mathrm{S(T)}$ denote the set of bipartitions (splits) induced by a phylogenetic tree $\mathrm{T}$. The RF distance between two trees $\mathrm{T}_1$ and $\mathrm{T}_2$ is defined as the number of bipartitions that are present in one tree but absent in the other, equivalently expressed as the symmetric difference between their split sets:

$$\mathrm{RF}(\mathrm{T}_1, \mathrm{T}_2) = |\mathrm{S}(\mathrm{T}_1) \setminus \mathrm{S}(\mathrm{T}_2)| + |\mathrm{S}(\mathrm{T}_2) \setminus \mathrm{S}(\mathrm{T}_1)| \quad (4)$$

where $\mathrm{S}(\mathrm{T}_1)$ and $\mathrm{S}(\mathrm{T}_2)$ denote the sets of bipartitions in the two trees, respectively. For two unrooted binary trees with an identical set of n taxa, the maximum possible RF distance is $2\mathrm{n} - 6$. The normalized Robinson–Foulds distance (nRF) was calculated as:

$$n\mathrm{RF} = \frac{\mathrm{RF}}{2n - 6} \quad (5)$$

The nRF value ranges from 0 (identical topologies) to 1 (completely different trees).

We quantified topological agreement using the normalized Split Agreement (nSA), defined as the proportion of shared non-trivial splits between the reference tree and the inferred tree:

$$n\mathrm{SA} = \frac{|S_1 \cap S_2|}{n - 3} \quad (6)$$

where $\mathrm{S}_1$ and $\mathrm{S}_2$ denote the sets of non-trivial bipartitions of the reference and inferred trees, respectively, and $\mathrm{n} - 3$ is the maximum number of internal splits in an unrooted binary tree. Higher nSA values indicate greater topological similarity.

## IV. EXPERIMENTS

### *A. Datasets*

The image data are obtained from Rove-Tree-11, a public insect image dataset of pinned rove beetle specimens [8]. The current study uses 13,887 segmented dorsal images annotated with species-level taxonomic labels and associated reference phylogenetic information. The data are aggregated to genus level, resulting in 44 genera across three subfamilies. Genus-level analysis is used because the number of species and images per genus is highly imbalanced, whereas genus-level aggregation provides a consistent unit for representation learning and tree reconstruction.

All images are padded to square canvases with a white background and resized to 518 x 518 pixels. To mitigate genus-level imbalance, geometric augmentation, including rotation, translation, and flipping, is applied to genera with fewer than 50 images until each genus reaches at least 50 unique samples. For each genus, 50 images are then sampled to construct a balanced base set. Further augmentation results in 28,600 images, corresponding to 650 images per genus. The augmented dataset is split into training, validation, and test subsets in a 7:2:1 ratio.

The text data consist of curated genus-level morphological descriptions. For representative images, preliminary descriptions were generated using GPT-5.2 [28] with a structured prompt template covering seven morphological domains: coloration, pronotum, elytra, legs, head and neck, eyes, and antennae. These fields were selected to match standard beetle morphological descriptors. The model-generated drafts were reviewed and corrected by trained personnel to ensure correspondence with the visible specimen morphology, and the corrected descriptions were further checked by domain specialists for consistency and biological validity. Fig. 2 shows an example of the morphology description format used to construct the text modality. The complete text bank is provided as supplementary material or in the project repository.

### *B. Experiment Settings*

The image encoder is initialized with the pretrained checkpoint dinov2_vitb14_reg4_pretrain. Input images are resized to 518*518 pixels. Text embeddings are extracted using BioClinical-ModernBERT-base and stored as a frozen genus-level text feature bank. The contrastive temperature is set to 0.07. AdamW is used as the optimizer, with learning rate $1*10^{-4}$, weight decay $1*10^{-4}$, and batch size 32. Training nis conducted on a single NVIDIA GeForce RTX 3090 GPU with 24 GB memory under CUDA 13.0.

For phylogenetic reconstruction, RevBayes uses the dnPhyloBrownianREML likelihood, a uniform topology prior, and NNI and SPR topology proposals. The MCMC analysis employs the first 1,000 burn-in generations and 500,000 sampling generations, and posterior trees are summarized as the maximum clade credibility tree. Detailed priors and proposal weights are provided in the supplementary material.

### *C. Baselines*

Two sets of comparisons are performed. First, several pretrained visual backbones are evaluated as feature extractors: DINO [12], DINOv2 [13], DINOv3 [29], BioCLIP [17], EVA-02 [14], BEiT v2 [15], and ConvNeXt V2 [16]. Second, ablation-style strategy comparisons are performed using the DINOv2 backbone: original DINOv2 features, image-only fine-tuning, and image-text fine-tuning. The same tree reconstruction and topology metrics are used across comparisons.

## V. Results

### A. Model Comparison

Table I compares the phylogenetic reconstruction performance of different visual backbones and includes the best median image-based result reported by Hunt and Pedersen [9]. Their approach uses an ImageNet-pretrained ResNet50 encoder, a 128-dimensional latent representation, and deep metric learning objectives to extract phylogenetically informative traits from insect specimen images. Among the evaluated models, DINOv2 achieves the lowest nRF and the highest nSA, indicating the closest topological agreement with the reference tree. DINOv2 is therefore selected as the main image encoder for further fine-tuning and multimodal alignment experiments.

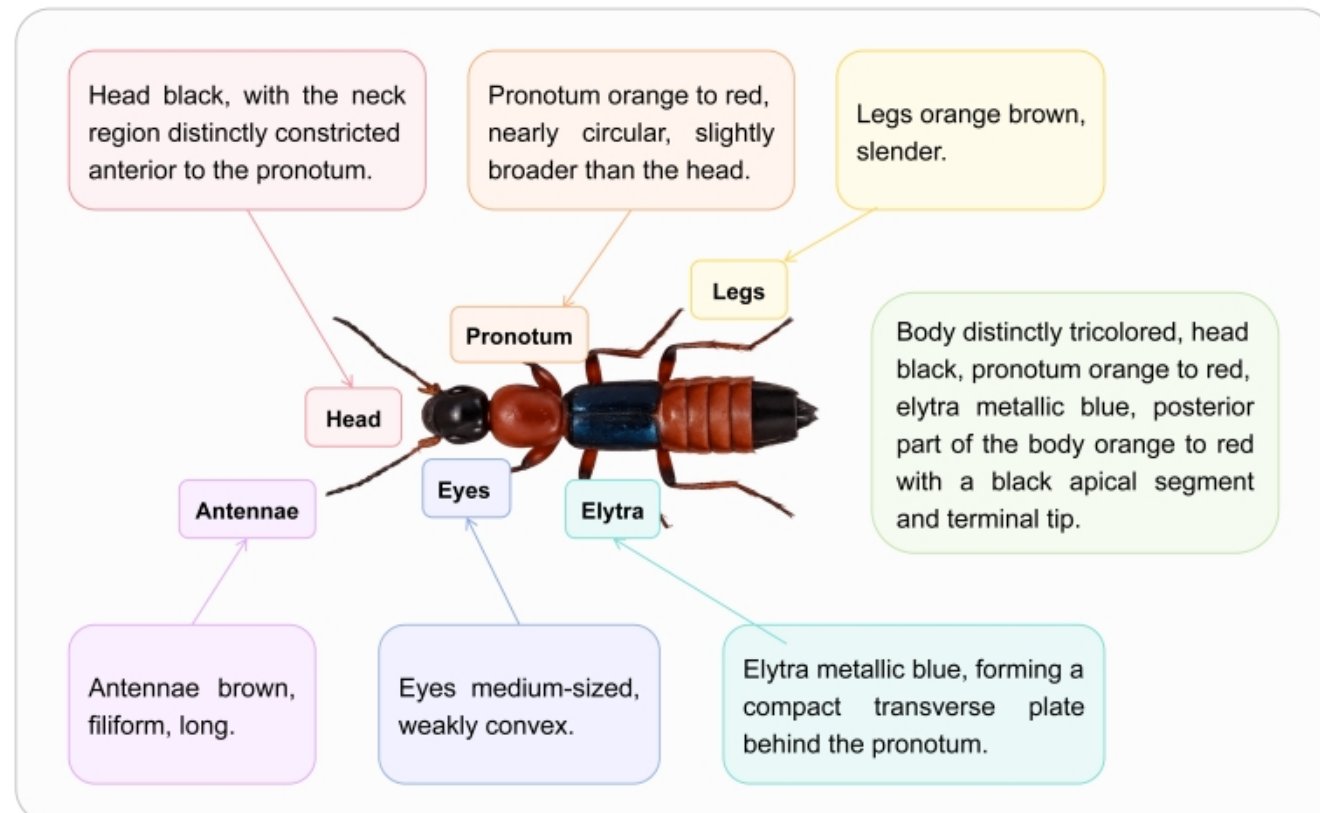

**Fig. 2.** Example genus-level morphological descriptions used to construct the text modality.

TABLE I
PERFORMANCE OF DIFFERENT VISUAL BACKBONES IN PHYLOGENETIC RECONSTRUCTION.

| Models | nRF | nSA |
|---|---|---|
| DINO | 0.8659 | 0.0488 |
| DINOv2 | 0.7195 | 0.1951 |
| DINOv3 | 0.7683 | 0.1463 |
| BioCLIP | 0.8659 | 0.0488 |
| EVA-02 | 0.8902 | 0.0244 |
| BEiT v2 | 0.9146 | 0.0000 |
| ConvNeXt V2 | 0.8902 | 0.0244 |

### B. Ablation Study

Table III compares the original DINOv2 representation, image-only fine-tuning, and image-text fine-tuning. Image-only fine-tuning improves over the original representation, suggesting that adaptation to the insect image dataset enhances the phylogenetic relevance of the embedding space. Image-text fine-tuning achieves the best performance, with the lowest nRF and highest nSA, showing that morphological text alignment provides useful semantic regularization for learning continuous traits.

Fig. 3 shows the training dynamics of the supervised contrastive loss and image-text alignment loss during joint image fine-tuning and image-text alignment. Both losses decrease rapidly during early training and gradually stabilize, indicating stable optimization. These two objectives jointly characterize the optimization of the proposed image fine-tuning and image-text alignment framework. The alignment loss acts as an additional regularization term, encouraging the formation of a structured embedding space without introducing obvious training instability.

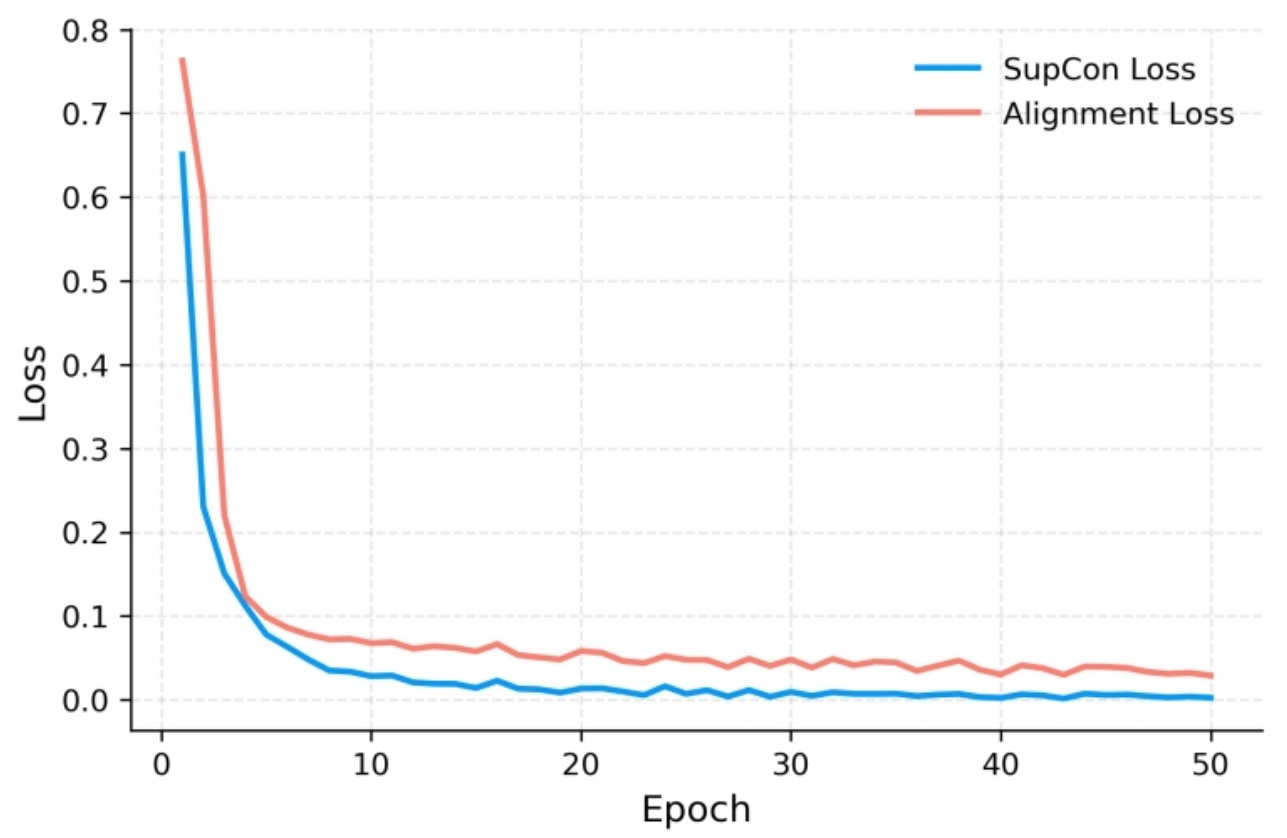

**Fig. 3.** Training dynamics of supervised contrastive loss and image-text alignment loss.

### C. Visualization of Attention Maps

Fig. 4 visualizes attention maps under three settings: the original DINOv2 model, image-only fine-tuning, and image-text fine-tuning. The original model produces broader attention patterns, whereas fine-tuning leads to more localized responses. After image-text alignment, attention remains concentrated on morphology-relevant regions, suggesting that the multimodal objective helps organize visual responses around discriminative anatomical structures.

TABLE II
ABLATION-STYLE COMPARISON OF RECONSTRUCTION PERFORMANCE.

| Backbone | Strategy | nRF | | nSA | |
|---|---|---|---|---|---|
| | | Average | Median | Average | Median |
| DINOv2 | Original | 0.705 ± 0.041 | 0.7195 | 0.210 ± 0.041 | 0.1951 |
| | Image fine-tuning | 0.6951 ± 0.000 | 0.6951 | 0.2195 ± 0.000 | 0.2195 |
| | Image-text fine-tuning | 0.6463 ± 0.000 | 0.6463 | 0.2683 ± 0.000 | 0.2683 |
| Hunt and Pedersen | Stratified dataset split | 0.829 ± 0.026 | 0.815 | - | - |

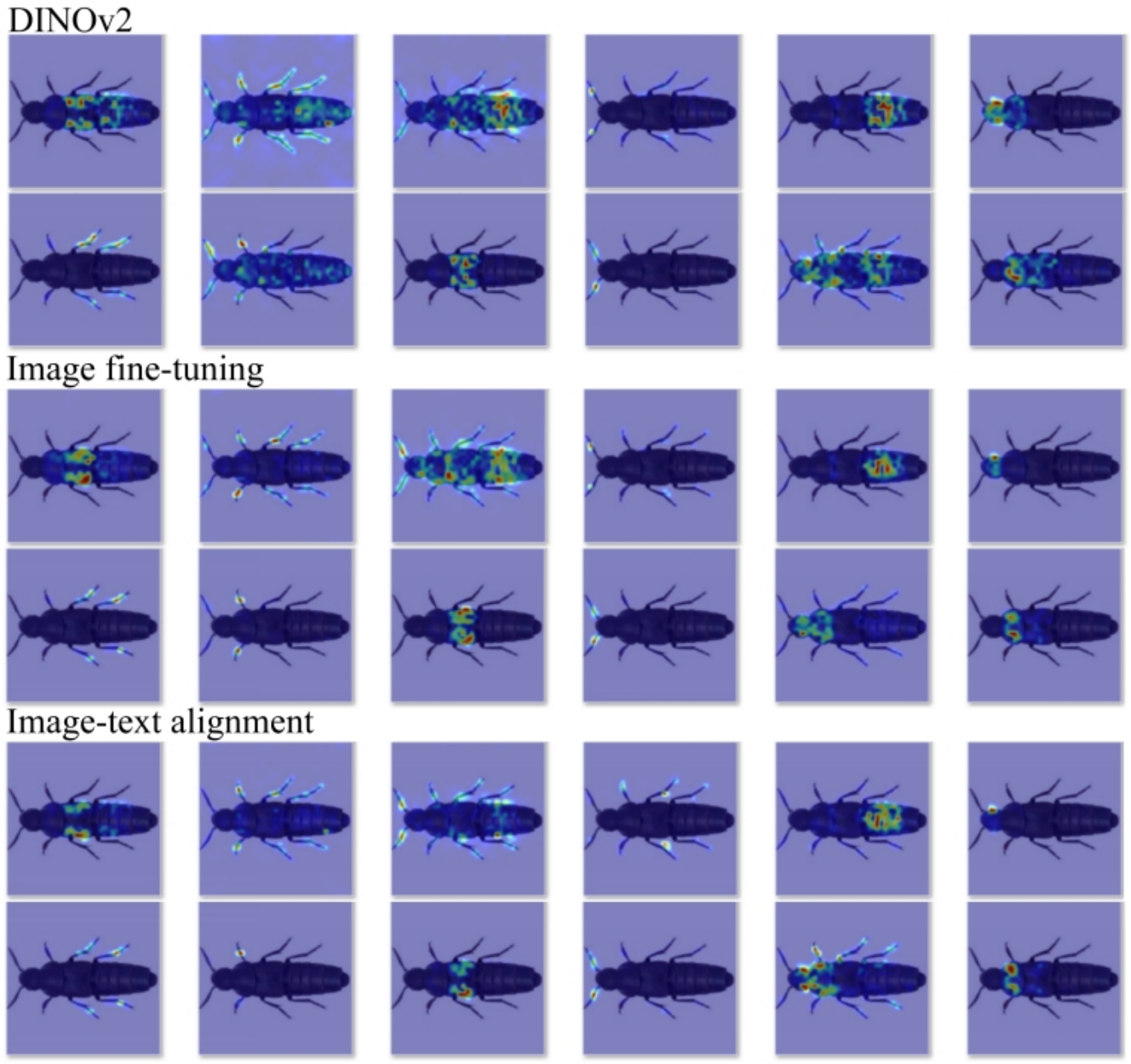

**Fig. 4.** Example of attention visualization of the Vision Transformer backbone on a test image.

### *D. Phylogenetic Analysis*

We compared the reference phylogeny with the tree inferred from multimodal features (Fig. 5.). The inferred topology recovers major subfamily-level groupings and shows improved agreement relative to baseline features. The quantitative results in Table II further support this observation: multimodal alignment reduces nRF and increases nSA compared with the original and image-only fine-tuned representations.

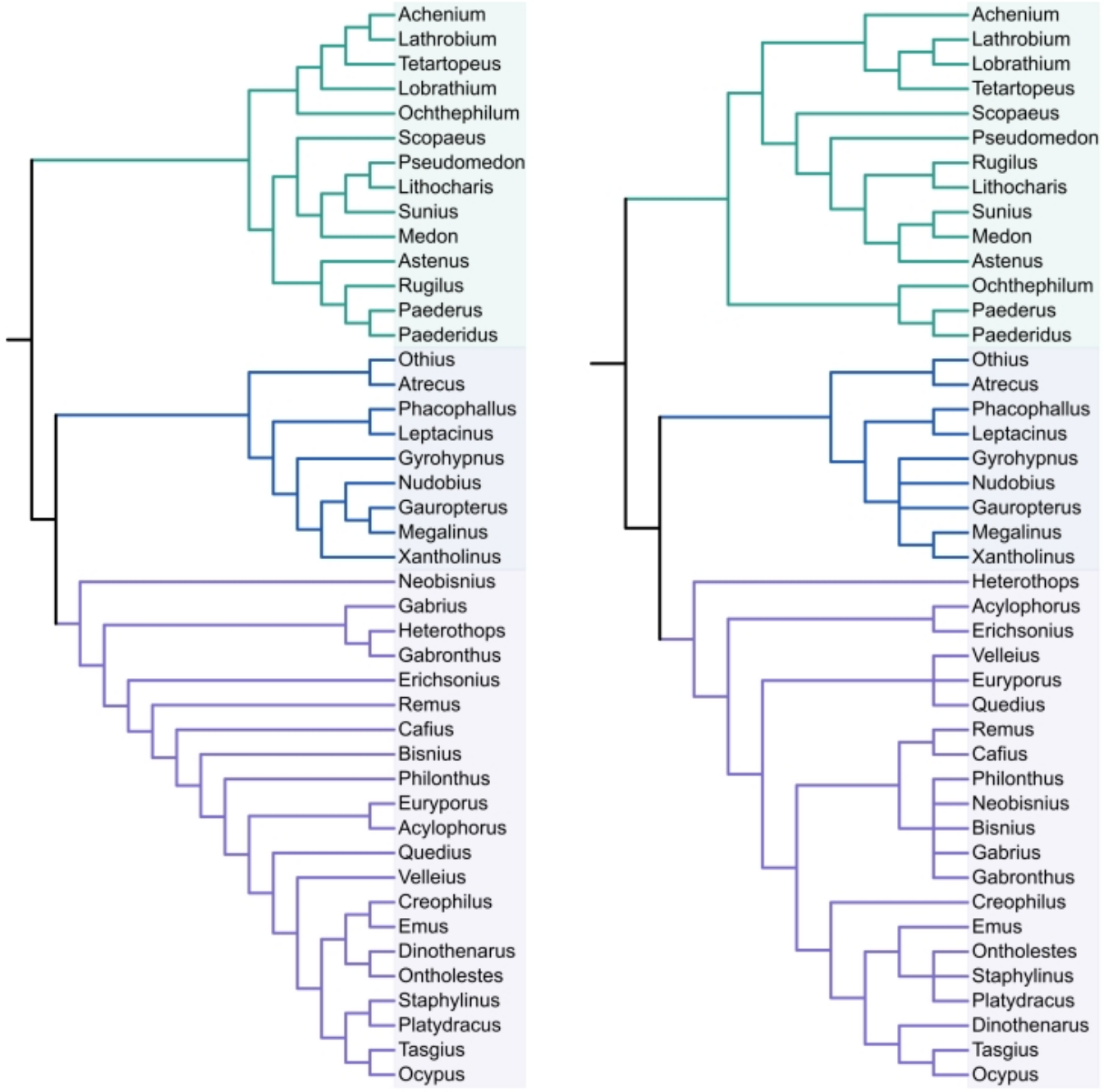

**Fig. 5.** Comparison of reference phylogeny and inferred tree reconstructed from multimodal features.

### *E. Discussion*

The multimodal morphology-based tree includes a local topological signal involving Gabronthus and Heterothops. In the reference phylogeny, these genera are separated by a larger evolutionary distance, whereas the inferred morphology-based tree places them closer together. To examine whether this pattern corresponds to a detectable molecular signal, a supplementary cytb-based analysis was performed. The cytb tree showed a similar local clustering pattern, while pairwise sequence identity and corrected maximum-likelihood distance did not support a globally close relationship.

Site-specific likelihood analysis identified two parallel amino-acid substitutions, Isoleucine at position 121 and Leucine at position 374, with relatively large local likelihood penalties. Such localized substitutions can contribute to apparent molecular convergence [30]. Phylogenetic analyses of convergent traits can also be sensitive to how shared changes are interpreted on a tree [31]. Related mitochondrial studies further illustrate the need to evaluate convergence carefully rather than treating every local similarity as evidence of close ancestry [32]. Gene-tree discordance provides one reason to treat this local pattern cautiously [33].

Difficult phylogenetic signals also require careful interpretation when a small number of sites exert strong influence [34]. In the supplementary molecular workflow, orthology inference used OrthoFinder [35]. Bootstrap approximation used UFBoot2 [36]. Model selection used ModelFinder [37]. Sequence alignment used MAFFT [38]. Maximum-likelihood tree inference used IQ-TREE 2 [39]. Evolutionary distance interpretation followed standard computational molecular evolution models [40]. This analysis is not used to revise the reference phylogeny; rather, it illustrates that morphology-aware multimodal representations can generate testable local signals for downstream molecular analysis. Full details of the molecular analysis are provided in the supplementary material.

## VI. Conclusion

This study presents a morphology-aware multimodal alignment framework for insect phylogenetic reconstruction. By adapting a self-supervised vision transformer with LoRA and supervised contrastive learning, and by aligning image embeddings with curated morphology descriptions, the framework derives continuous visual traits for Bayesian phylogenetic reconstruction. Experiments on Rove-Tree-11 show that image-text alignment improves topological agreement with the reference phylogeny compared with original and image-only fine-tuned representations. Attention visualization further indicates that multimodal alignment promotes morphology-relevant visual responses. Future work will evaluate the framework on broader datasets, incorporate uncertainty-aware embeddings, and investigate more expressive phylogenetic models for learned continuous traits.